# A Neural Network Training Method Based on Neuron Connection Coefficient Adjustments

Jiang Kun
Chongqing university
atan_j@qq.com

**Abstract**: In previous studies, we introduced a neural network framework based on symmetric differential equations, along with one of its training methods. In this article, we present another training approach for this neural network. This method leverages backward signal propagation and eliminates reliance on the traditional chain derivative rule, offering a high degree of biological interpretability. Unlike the previously introduced method, this approach does not require adjustments to the fixed points of the differential equations. Instead, it focuses solely on modifying the connection coefficients between neurons, closely resembling the training process of traditional multilayer perceptron (MLP) networks. By adopting a suitable adjustment strategy, this method effectively avoids certain potential local minima. To validate this approach, we tested it on the MNIST dataset and achieved promising results. Through further analysis, we identified certain limitations of the current neural network architecture and proposed measures for improvement.

## 1. Introduction

Since the early 20th century, researchers have systematically studied nerve cells, revealing that the signal propagation in neurons depends on changes in membrane potential. This process can be mathematically described by a set of differential equations[1]. However, due to the complexity of these equations, researchers have simplified them to emulate the functions of nerve cells. In this simplification, neurons are represented as numerical values, with their connection coefficients determining the output of these values[2]. This approach has led to significant advancements in neural network applications and forms the foundation of current mainstream large-scale neural networks.

Despite the remarkable success of mathematically based neural networks, there remains a strong interest in developing neural network systems inspired by biological cells. This interest arises from the fact that biological systems are fundamentally composed of cells, and the efficiency of biological nervous systems far exceeds that of mathematical neural networks. Consequently, constructing neural networks modeled on biological systems offers not only superior biological interpretability but also considerable potential for efficiency improvements. Existing biological neural networks, such as cellular neural networks, chaotic neural networks, and spiking neural networks, have demonstrated notable achievements[3-5]. However, their adoption remains limited, primarily because biological neural networks lack the versatility of their mathematical counterparts.

Therefore, we propose a neural network architecture based on symmetric differential equations, offering both high operability and strong biological interpretability[6]. In this approach, symmetric differential equations replace the numeral neurons in the traditional MLP, forming a novel network structure. To investigate this new structure, we first focus on studying a single system. Specifically, we model the neuron as a system governed by a set of differential equations. We analyze the fixed points of this system, define the signal propagation mechanism, and ultimately propose a novel neural network training method. These foundational studies have been detailed in previous work[7].

In previous research, we introduced the first training method for this neural network framework, which achieved output modification by altering the system's fixed points. This approach to fixed point adjustment is a distinctive method rooted in chaos theory, making it unique to neural networks based on differential equations. In contrast, contemporary large-scale mathematical neural networks typically focus on adjusting the connection coefficients between neurons. In this study, we shift away from fixed point adjustment and instead train the neural network by modifying the connection coefficients between neurons, aligning more closely with the traditional MLP training paradigm[8]. Under this second training method, the differential equations serve the roles of activation and summation functions, while their parameters of differential equations remain unchanged throughout the training process.

In the field of neural network training, the backpropagation algorithm has been the dominant method, enabling mathematical neural networks to achieve significant success while leaving biological neural networks without a high effective training methods[5, 9]. The training method proposed in this paper builds upon the principles of backpropagation but departs from the traditional implementation, supporting our earlier assertion that the backward signal propagation can serve as an alternative to the chain derivative method[7]. In previous studies, the coefficients between neurons were fixed to adhere to the principle of instinctive design, which states that the operation of neurons should not extend beyond their intrinsic range. However, this constraint limited the scope of adjustments available to neurons. In this paper, we expand the scope of neural control by allowing neurons to communicate with adjacent neurons and adjust the connection coefficients between them. This expansion, however, remains local—neurons are only required to process signals from their immediate neighbors.

A major challenge for neural networks based on differential equations is the computational intensity of solving such equations. Despite the availability of alternative and optimized methods, this challenge continues to impede the large-scale deployment of biological neural networks[10, 11]. In our current research, while computer simulations are used to iteratively calculate the states of neural networks over time, the system design is inherently hardware-oriented[12]. Each neuron only processes local signals, enabling the network to scale effectively through direct hardware connections. Differential equations, as mathematical constructs, can be directly implemented through hardware circuits. Consequently, once the system is realized at the hardware level, the dynamic evolution of the differential equations can occur autonomously, eliminating the need for additional computational resources. This not only enhances the system's efficiency but also improves its biological interpretability.

The following chapters of this paper are arranged as follows. Chapter 2 will briefly introduce the Wuxing neural network, mainly to let everyone understand the basic framework of the network and the connection coefficients we adjusted. In our previous paper, we provided a more detailed introduction. Chapter 3 will introduce the adjustment method of connection coefficients and the study of neural network structure. The last chapter is the summary and future prospects.

## 2. Wuxing neural network

For a long time, mathematically based neural networks have dominated the field of neural network development, even though biological nerve cells are far more complex than simple numerical representations. To enhance the biological interpretability of neural networks, many biologically inspired neural network models have been proposed. However, the training of these networks either relies on traditional backpropagation algorithms or employs inefficient strategies. In the first case, the biological neural networks sacrifice interpretability, conflicting with the goal of designing biologically plausible systems. In the second case, inefficiency limits the scalability and general applicability of these networks.

To address this issue, we propose a neural network framework grounded in symmetric differential equations. By integrating symmetry with differential equations, this framework achieves both favorable mathematical properties and strong operability. In this framework, continuous derivative operations are no longer required, as they can be replaced by backward signal propagation. Backward signal propagation is entirely reasonable and feasible at the scale of biological cells. Consequently, our approach combines the biological interpretability of biologically inspired systems with the efficiency and scalability of mathematical neural networks, making it suitable for implementation at the biological level.

So far, our discussion of this theoretical framework remains partial, we believe that as more methods and theories are introduced, it will become evident that this is a powerful and versatile neural network framework. In prior studies, we introduced the Wuxing neural network based on the Wuxing theory. Although a general expression for the network was provided, this paper focuses on the Wuxing neural network to maintain a clear scope of discussion. The expression for the Wuxing neural network is shown in Equation 2.1, where $E$ represents different elements, $K_1$, $K_2$ and $K_3$ are three sets of parameters, and the superscripted numbers denote offsets following a fixed sequence.

$$\frac{d\overset{0}{E}}{dt} = K_1 \overset{0}{E}\overset{-1}{} - K_2 \overset{0}{E}\overset{0}{} - K_3 \overset{0}{E}\overset{0}{}\overset{-2}{E} \tag{2.1}$$

It is a system with a specific fixed point. When the parameters in $K_1$, $K_2$ and $K_3$ are equal, the fixed point $B_0$ can be obtained according to equation 2.2.

$$B_0 = \frac{K_1 - K_2}{K_3} \tag{2.2}$$

In this system, we consider the state of the system at its fixed point as its zero state. When the system receives external signals, it deviates from the fixed point. These deviations generate new signals, which are continuously propagated through the connections between neurons, eventually forming signals with various logical structures. This process is analogous to throwing stones into still water, where the resulting ripples spread outward continuously.

$$\frac{d\overset{0}{E(t)}}{dt} = K_1 \overset{0}{E(t)}\overset{-1}{} - K_2 \overset{0}{E(t)}\overset{0}{} - K_3 \overset{0}{E(t)}\overset{0}{}\overset{-2}{E(t)} + Input(t) \tag{2.3}$$

In equation 2.3, $Input(t)$ is the input signal. According to our settings, the output signal $D(t)$ can be determined by formula 2.3.

$$D(t) = E(t) - B_0 \tag{2.4}$$

In equation 2.4, $E(t)$ is the element value that changes with time, $B_0$ is the fixed point of the system, and the signal generated by the system is actually the value that deviates from the fixed point.

By design, the system is reversible, allowing error signals to propagate backward. This reversibility significantly enhances the trainability of the system, as it eliminates the need to project the system onto an external framework for training. Unlike methods such as particle swarm optimization or genetic algorithms, this approach relies solely on the system's inherent signal propagation. Furthermore, it bypasses the need for continuous partial derivatives, streamlining the training process while maintaining the system's internal coherence.

According to the definition of reversibility of the system, in the reverse system, the causal relationship between elements changes, but the connection relationship does not change. We can get the system equation of backward propagation.

$$\frac{d \overset{0}{E}(t)}{dt} = \overset{1}{K_1} \overset{1}{E}(t) - \overset{0}{K_2} \overset{0}{E}(t) - \overset{2}{K_3} \overset{0}{E}(t) \overset{2}{E}(t) + Input(t) \tag{2.5}$$

In equation 2.5, the order of element $E$ and parameters $K_1$, $K_2$ and $K_3$ are adjusted. This is because we only changed the causal order of the elements without changing their connection relationship. Similarly, we can also get the signal $\hat{D}(t)$ in the backward propagation network.

$$\hat{D}(t) = \hat{E}(t) - \hat{B}_0 \tag{2.6}$$

where $\hat{E}(t)$ is the element value in the backward propagation and $\hat{B}_0$ is the fixed point determined by the backward propagation Equation (2.6).

## 3. Connection coefficient adjustment

In the neural network framework we propose, neurons constructed from differential equations do not require additional activation functions, as the system is inherently nonlinear and bounded. In contrast, traditional neural networks lack inherent nonlinearity and therefore rely on the explicit introduction of activation functions. In traditional multi-layer perceptron (MLP) neural networks, the output is typically adjusted by modifying the connection coefficients between neurons[8]. Similarly, in this paper, we control network output by tuning the connection coefficients between neurons.

It is important to note that the training method used in this study represents the second neural network training approach, which is fully compatible with the first method. In previous studies, we trained neural networks by adjusting the internal parameters of neurons to modify the system's equilibrium points. This method is fundamentally different from the approach employed in this paper. However, the two methods are not mutually exclusive, allowing for the simultaneous use of both approaches to enhance the training process.

Figure 1 shows a simple neural network, where Figure 1a is the forward propagation network and Figure 1b is the backward propagation network. There are different connection coefficients $C = \{c_{ij}\}$, between different neurons, so the forward propagation equation is modified to:

$$\frac{d \overset{0}{E}(t)}{dt} = \overset{0}{K_1} \overset{-1}{E}(t) - \overset{0}{K_2} \overset{0}{E}(t) - \overset{0}{K_3} \overset{0}{E}(t) \overset{-2}{E}(t) + C \bullet Input(t) \tag{3.1}$$

The backward propagation equation is modified as follows:

$$\frac{d \overset{0}{E}(t)}{dt} = \overset{1}{K_1} \overset{1}{E}(t) - \overset{0}{K_2} \overset{0}{E}(t) - \overset{2}{K_3} \overset{0}{E}(t) \overset{2}{E}(t) + C \bullet Input(t) \tag{3.2}$$

In equation 3.1 and 3.2, the coefficients $C$ remains consistent across corresponding positions, ensuring causal consistency before and after connections. To maintain a one-to-one correspondence between cause and effect, each neuron's synaptic connections are strictly one-to-one. Compared with the fully connected structure of traditional MLPs, the connections in the Wuxing neural network are significantly sparser.

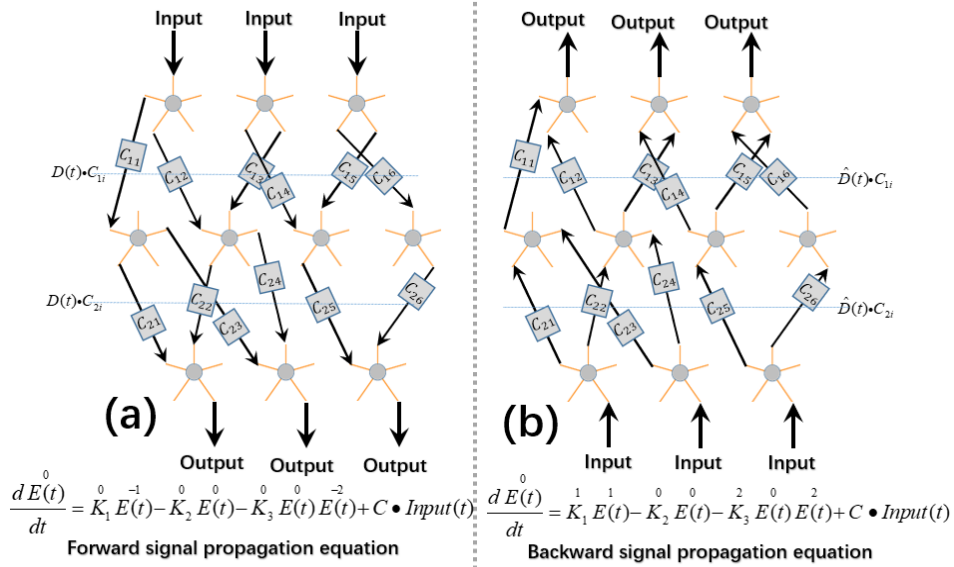

**Fig.1 The forward and backward propagation of Wuxing neural networks**

Fig. 1. a Forward propagation path of Wuxing neural network and corresponding equations. This network adopts a randomly connected multi-layer neural network structure. Unlike the traditional MLP, each neuron has only 5 interfaces, and the connections between the interfaces are one-to-one.

Fig. 1 b Backward propagation path of Wuxing neural network and corresponding equations. Compared with the forward network structure, the connection method has not changed, the only change is the direction of signal propagation. The backward propagation equation is changed according to the connection method.

To quantify the connectivity between layers, we define the link ratio (LR) as the ratio of the actual number of connections to the total possible connections in a fully connected network. For example, in Figure 1a, the first layer contains 3 neurons, and the second layer contains 4 neurons. In a fully connected network, there would be 12 connections between these two layers. However, in this case, there are only 6 connections, resulting in a link ratio (LR) of 50% from the first layer to the second layer.

As a verification, we can use the previously established model MODEL1, the model has 784 inputs, 10 outputs, and a total of 6 layers. The number of neurons in each layer is: {784, 839, 283, 96, 32, 10}, of which the first and last layers are fully connected, and all interfaces have inputs or outputs. The initial parameters of the model are $K_1$={1, 1, 1, 1, 1}; $K_2$={0.5, 0.5, 0.5, 0.5, 0.5}; and $K_3$={0.5, 0.5, 0.5, 0.5, 0.5}.

Building on our previous research, we continue to employ the training method based on backward signal propagation (BSP). The BSP method operates on the premise that the system is reversible. In such a reversible system, input signals from both ends are projected through the network structure and connection coefficients. By comparing the differences in signals between the two propagation directions, the corresponding adjustments can be determined. Our verification demonstrates that the BSP method can effectively replace the traditional back-propagation algorithm. Furthermore, it is distributed and possesses high biological interpretability. For a single neuron, the information required is local, while global information is achieved through the propagation of signals across the network.

According to our research, we assume that the system has n layers, where the forward output signal of the i-th layer neuron is $D_i(t)$, and similarly, the backward output signal of the i-th layer neuron is $\hat{D}_i(t)$. If the calculation time is $T$, we can define a variable $G_1$

$$G_1 = \int_0^T D_i(t)dt \bullet \int_0^T \hat{D}_{i+1}(t)dt \tag{3.3}$$

G1 represents the causal quantity of the connection parameters between the neurons in the i-th layer and the neurons in the i+1-th layer.

Since $G_1$ may exceed a certain limit, we use the inverse tangent function (other similar functions are also possible) to limit it and get $G_2$

$$G_2 = atan(G_1 * kt) / kt \tag{3.4}$$

$kt$ is the adjustment parameter, $G_2$ is the adjusted correlation value. The parameter can be adjusted based on $G_2$.

$$C_{new} = C_{old} + G_2 \tag{3.5}$$

In equation 3.5, we employ addition instead of multiplication, a key distinction from the first training method. This modification helps the system effectively avoid local minima. To ensure system stability, we also constrain the range of C. In our prior studies, we introduced a distributed Proportional-Integral-Derivative (PID) control method but found that the differential control method lacked stability. To address this, we proposed the development of an improved differential control method. The approach introduced in this article serves as such a method. Since the coefficients adjustments in this new method do not conflict with the previous training approach, both the first and second training methods can be utilized simultaneously.

### Fig. 2 Accuracy curves on MODEL 1

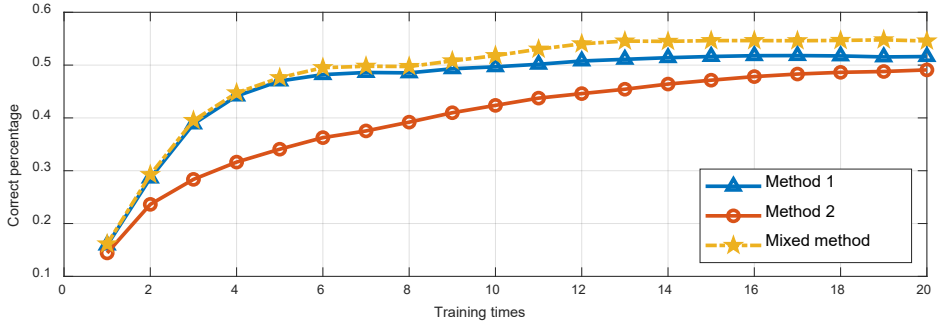

Fig. 2 The accuracy curves of MODEL1 trained using the first method, the second method, and the mixed method. The results show that both the second method and the first method can effectively train the neural network, and if the two methods are mixed, the accuracy can be further improved.

Figure 2 presents the accuracy curves of MODEL1 trained using the first method, the second method, and the mixed method. It is not meaningful to compare the training speeds of methods 1 and 2 in Figure 2, as they utilize different adjustment parameters. To ensure system stability, the adjustment step for method 2 is limited to only 25% of that for method 1. Nevertheless, the results demonstrate that both method 1 and method 2 can effectively enhance the system's accuracy.

During the training of MODEL1, regardless of the method employed, the accuracy eventually plateaus, unable to improve further. In a previous study, we noted that system stability is maintained by constraining the system parameters within a specific range. However, the primary limitation in MODEL1 is not the parameter range but the link ratio LR. As previously defined, the average link ratio LR between layers in MODEL1 is just 0.51%. This is due to the fact that each neuron has only 5 connections, resulting in a very low link ratio LR between layers.

**Fig. 3 Accuracy curves on MLP based on MODEL 1 of different link ratio LR**

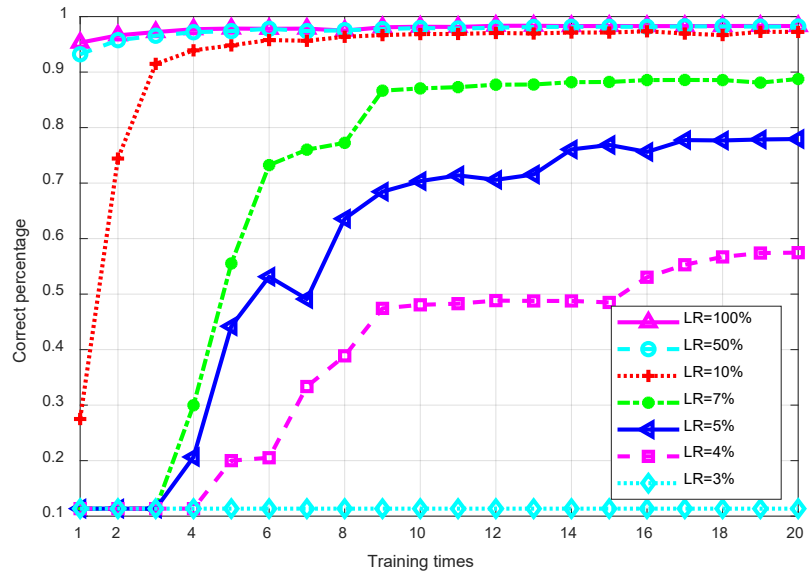

Fig. 3 This is the training accuracy curve of the MLP model based on MODEL1 under different link ratio LR. When LR is relatively large, the model can learn well, but when LR is less than 10%, the learning ability of the model begins to drop sharply.

As a validation step, we constructed an MLP model with the same number of layers and neurons as MODEL1 but varied the link ratio LR between each layer. This allowed us to obtain the training accuracy curves under different LR values (Figure 3). As shown in Figure 3, when LR is relatively high, the model can be trained quickly to achieve very high accuracy. However, as LR decreases below 10%, training the model becomes increasingly challenging. In particular, when LR falls below 5%, the difficulty of training rises sharply, and according to our tests, it becomes almost impossible to train the model effectively when LR is less than 3%.

In the Wuxing neural network MODEL1, the link ratio LR is only 0.51%, which imposes a significant limitation on the model's learning capacity. Nevertheless, this also highlights the robustness of our model, as it can still capture a substantial amount of system features even with such a low connection ratio.

Addressing the issue of an excessively low link ratio LR in the Wuxing neural network is relatively straightforward. One solution is to incorporate an adder in front of the synapse of each neuron, allowing a single neuron synapse to receive signals from multiple neurons while still adhering to our propagation rules. With this adjustment, the neural network framework we designed achieves full integration with modern neural network frameworks.

Additionally, there are alternative ways to enhance the learning capacity of the neural network, such as increasing the number of neurons. In MODEL1, we did not implement adders; however, we can adjust the number of neurons in the middle layers to create MODEL2. In this configuration, the number of neurons in each layer is {784, 1048, 353, 119, 40, 10}.

In MODEL2, the system's link ratio (LR) is 0.38%, yet its learning ability surpasses that of MODEL1. This indicates that some connections in MODEL1 are redundant or ineffective. Consequently, adopting an optimized connection strategy can enhance the system's learning ability, which is why MODEL2 is highlighted here.

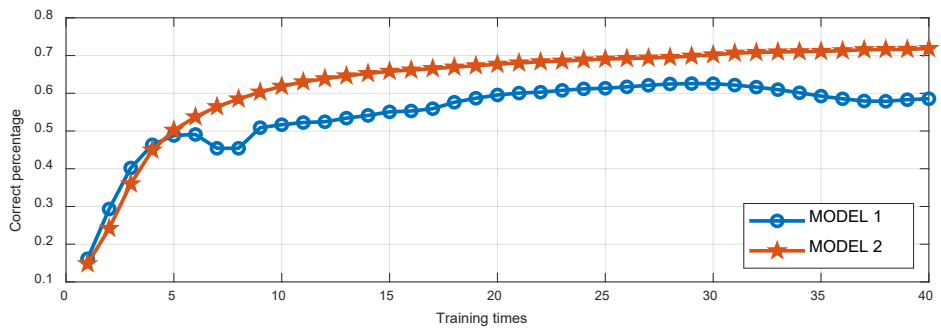

**Fig. 4 Accuracy curves on MODEL 1 and MODEL 2**

Fig. 4 We use a wide range of parameters to train MODEL 1 and MODEL 2 for a longer time, and we can see that both models can improve the accuracy. However, since MODEL 2 has a stronger learning ability, the training effect on MODEL 2 is more satisfactory.

## 4. Summary

In this paper, we utilize the previously proposed Wuxing neural network to establish a neural network that more closely resembles a Multilayer Perceptron (MLP). Within this network, we have incorporated connection coefficients between neurons for training, which bears a strong resemblance to traditional MLP architectures. The results indicate that independently adjusting the connection coefficients between neurons can also facilitate effective learning, and is compatible with the first training method proposed earlier. Since the adjustment in this method occurs before the neurons, it can function as the differential component in PID control.

In our research, different training methods can be effectively integrated. Therefore, we combined the first and second methods and tested them on MODEL1. The results demonstrate that the mixed method possesses superior learning capabilities. However, further analysis reveals that the primary constraint on the system's learning ability is the number of connections between neurons. We propose a solution to increase the number of neuron connections by adding adders at the neuron synapses, which effectively enhances the connectivity. Additionally, we investigated the method of improving the system's learning ability by increasing the number of neurons. The results show that appropriately increasing the number of neurons, leaving some neuron synapses unconnected, enhances the system's learning capacity. However, these are merely superficial observations; it is important to note that effective and properly regulated neuron connection methods contribute significantly to improving the system's learning ability.

Currently, our research on the entire system is still partial. Under various constraints, the neural network has still achieved effective learning outcomes. We believe that learning under constraints can better demonstrate the effectiveness of our method. In this paper, we achieved highly efficient learning with an extremely low neuron connection ratio (LR), indicating that our proposed neural network has excellent learning capabilities. Therefore, we will continue to introduce new training methods under constraints and integrate them with existing methods to achieve better learning outcomes. This restriction will not only effectively validate the method's efficacy but also accurately reflect the system's learning capacity.

## Acknowledgements

Thanks to China Scholarship Council (CSC) for their support during the pandemic, which allowed me to get through those difficult days and give me the opportunity to put my past ideas into practice, ultimately resulting in the article I am sharing with you today.

# Reference

[1] A. L. Hodgkin and A. F. Huxley, "A quantitative description of membrane current and its application to conduction and excitation in nerve," *The Journal of physiology,* vol. 117, no. 4, p. 500, 1952.

[2] K. Hornik, M. Stinchcombe, and H. White, "Multilayer feedforward networks are universal approximators," *Neural networks,* vol. 2, no. 5, pp. 359-366, 1989.

[3] L. O. Chua and L. Yang, "Cellular neural networks: Theory," *IEEE Transactions on circuits and systems,* vol. 35, no. 10, pp. 1257-1272, 1988.

[4] K. Aihara, T. Takabe, and M. Toyoda, "Chaotic neural networks," *Physics letters A,* vol. 144, no. 6-7, pp. 333-340, 1990.

[5] A. Tavanaei, M. Ghodrati, S. R. Kheradpisheh, T. Masquelier, and A. Maida, "Deep learning in spiking neural networks," *Neural networks,* vol. 111, pp. 47-63, 2019.

[6] K. Jiang, "A Neural Network Framework Based on Symmetric Differential Equations," doi: http://dx.doi.org/10.12074/202410.00055.

[7] K. Jiang, "A Neural Network Training Method Based on Distributed PID Control," *arXiv preprint arXiv:2411.14468,* doi: https://doi.org/10.48550/arXiv.2411.14468.

[8] L. Noriega, *Multilayer perceptron tutorial* (School of Computing, no. 5). Staffordshire University, 2005, p. 444.

[9] D. E. Rumelhart, R. Durbin, R. Golden, and Y. Chauvin, "Backpropagation: The basic theory," in *Backpropagation*: Psychology Press, 2013, pp. 1-34.

[10] R. Hasani, M. Lechner, A. Amini, D. Rus, and R. Grosu, "Liquid time-constant networks," in *Proceedings of the AAAI Conference on Artificial Intelligence*, 2021, vol. 35, no. 9, pp. 7657-7666.

[11] R. Hasani *et al.*, "Closed-form continuous-time neural networks," *Nature Machine Intelligence,* vol. 4, no. 11, pp. 992-1003, 2022.

[12] S. P. Adhikari, C. Yang, H. Kim, and L. O. Chua, "Memristor bridge synapse-based neural network and its learning," *IEEE Transactions on neural networks and learning systems,* vol. 23, no. 9, pp. 1426-1435, 2012.